\documentclass[nohyperref]{article}

\usepackage{microtype}
\usepackage[pdftex]{graphicx}
\usepackage{subfigure}
\usepackage{booktabs} 
\usepackage{adjustbox}

\usepackage[utf8]{inputenc} 
\usepackage[T1]{fontenc}    
\usepackage[pagebackref,breaklinks,colorlinks]{hyperref}
\hypersetup{
    colorlinks=true,
    citecolor=cobalt,
    linkcolor=Red,
    urlcolor=Green,
    }

\usepackage{amsmath}
\usepackage{amssymb}
\usepackage{booktabs}
\usepackage[utf8]{inputenc} 

\usepackage{adjustbox}

\usepackage[nameinlink]{cleveref}
\crefname{section}{Sec.}{Secs.}
\Crefname{section}{Section}{Sections}
\crefformat{section}{\S#2#1#3}

\crefname{table}{Tab.}{Tabs.}
\Crefname{figure}{Fig.}{Figs.}

\usepackage{textcomp}
\usepackage[dvipsnames, table]{xcolor}
\usepackage{MnSymbol,bbding,pifont}

\usepackage{microtype} 

\usepackage{amsmath, amssymb, amsthm}
\usepackage{mathtools}
\usepackage{bbm}
\usepackage{dsfont}
\usepackage{thmtools}
\usepackage{thm-restate}

\usepackage{graphicx, wrapfig}

\usepackage{booktabs, array}
\usepackage{multirow}
\usepackage{adjustbox}

\usepackage{textcomp}

\usepackage{adjustbox}
\usepackage[T1]{fontenc}

\usepackage[algoruled]{algorithm2e}
\usepackage{listings}
\usepackage{fancyvrb}
\fvset{fontsize=\small}
\usepackage[dvipsnames, table]{xcolor}

\usepackage{MnSymbol,bbding,pifont}

\definecolor{turquoise}{cmyk}{0.65,0,0.1,0.3}
\definecolor{purple}{rgb}{0.65,0,0.65}
\definecolor{dark_green}{rgb}{0, 0.5, 0}
\definecolor{orange}{rgb}{0.8, 0.6, 0.2}
\definecolor{red}{rgb}{0.8, 0.2, 0.2}
\definecolor{darkred}{rgb}{0.6, 0.1, 0.05}
\definecolor{blueish}{rgb}{0.0, 0.3, .6}
\definecolor{light_gray}{rgb}{0.7, 0.7, .7}
\definecolor{pink}{rgb}{1, 0, 1}
\definecolor{greyblue}{rgb}{0.25, 0.25, 1}
\definecolor{pastelgreen}{rgb}{0.47, 0.87, 0.47}
\definecolor{teagreen}{rgb}{0.82, 0.94, 0.75}
\definecolor{cobalt}{rgb}{0.0, 0.28, 0.67}

\newcommand{\loss}[1]{\mathcal{L}_\text{#1}}

\usepackage{xspace}
\newcommand*{\eg}{\emph{e.g.}\@\xspace}
\newcommand*{\ie}{\emph{i.e.}\@\xspace}

\usepackage{blindtext}

\newcommand{\norm}[1]{\left\lVert#1\right\rVert}

\theoremstyle{plain}

\theoremstyle{definition}

\theoremstyle{remark}

\PassOptionsToPackage{square}{natbib} 
\usepackage[preprint]{neurips_2023}
\setcitestyle{numbers} 

\usepackage[utf8]{inputenc} 
\usepackage[T1]{fontenc}    
\usepackage{hyperref}       
\usepackage{url}            
\usepackage{booktabs}       
\usepackage{amsfonts}       
\usepackage{nicefrac}       
\usepackage{microtype}      

\title{Visualizing the loss landscape of Self-supervised Vision Transformer}

\newcommand*{\affaddr}[1]{#1} 
\newcommand*{\affmark}[1][*]{\textsuperscript{#1}}

\author{%
\affmark[1,2]Youngwan Lee~~~\affmark[2]Jeffrey Willette~~~~\affmark[1]Jonghee Kim~~~~\affmark[2]Sung Ju Hwang\\ 
\vspace{-0.1in}
\\
\affaddr{\affmark[1]Electronics and Telecommunications Research Institute~(ETRI), South Korea}\\
\affaddr{\affmark[2]Korea Advanced Institute of Science and Technology~(KAIST), South Korea}\\
}

\begin{document}

\maketitle

\begin{abstract}

The Masked autoencoder (MAE) has drawn attention as a representative self-supervised approach for masked image modeling with vision transformers. However, even though MAE shows better generalization capability than fully supervised training from scratch, the reason why has not been explored.
In another line of work, the Reconstruction Consistent Masked Auto Encoder (RC-MAE), has been proposed which adopts a self-distillation scheme in the form of an exponential moving average (EMA) teacher into MAE, and it has been shown that the EMA-teacher performs a conditional gradient correction during optimization. To further investigate the reason for better generalization of the self-supervised ViT when trained by MAE (MAE-ViT) and the effect of the gradient correction of RC-MAE from the perspective of optimization, we visualize the loss landscapes of the self-supervised vision transformer by both MAE and RC-MAE and compare them with the supervised ViT (Sup-ViT). Unlike previous loss landscape visualizations of neural networks based on classification task loss, we visualize the loss landscape of ViT by computing pre-training task loss. 
Through the lens of loss landscapes, we find two interesting observations: (1) MAE-ViT has a smoother and wider overall loss curvature than Sup-ViT. (2) The EMA-teacher allows MAE to widen the region of convexity in both pretraining and linear probing, leading to quicker convergence.
To the best of our knowledge, this work is the first to investigate the self-supervised ViT through the lens of the loss landscape.

\end{abstract}

\section{Introduction}\label{sec:intro}

Due to the scalability and versatility of self-attention mechanisms~\cite{vaswani2017attention}, the Vision Transformer~(ViT)~\cite{dosovitskiy2021vit} has been widely used in the vision domain from image/pixel-level recognition~\cite{liu2021swin,Fan_2021mvit,lee2022mpvit,el2021xcit} to video applications~\cite{feichtenhofer2022mae_video,tong2022videomae,bertasius2021timesformer}.
In the recent self-supervised learning literature~\cite{bao2021beit,he2022mae,zhou2021ibot,assran2022msn,dong2022bootmae,lee2023rcmae}, masked image modeling such as the Masked autoencoder~\cite{he2022mae}~(MAE) which utilizes a ViT backbone and predicts masked patches given a set of unmasked patches, has been a staple pre-text task. MAE achieves better generalization performance than a fully supervised ViT from scratch,
but, the reasons for the superior generalization have not been widely explored yet.

Meanwhile, Lee et al.~\cite{lee2023rcmae} have proposed the reconstruction-consistent Masked autoencoder~(RC-MAE) by adopting a self-distillation scheme~\cite{grill2020byol,caron2021dino} in the form of an exponential moving average (EMA) teacher into the MAE framework. The RC-MAE showed that the EMA-teacher performs conditional gradient corrections during optimization.
As a result, RC-MAE showed better generalization performance on downstream tasks and achieved a faster convergence speed than MAE.

\begin{figure}[t]
\begin{center}
\includegraphics[width=0.8\textwidth]{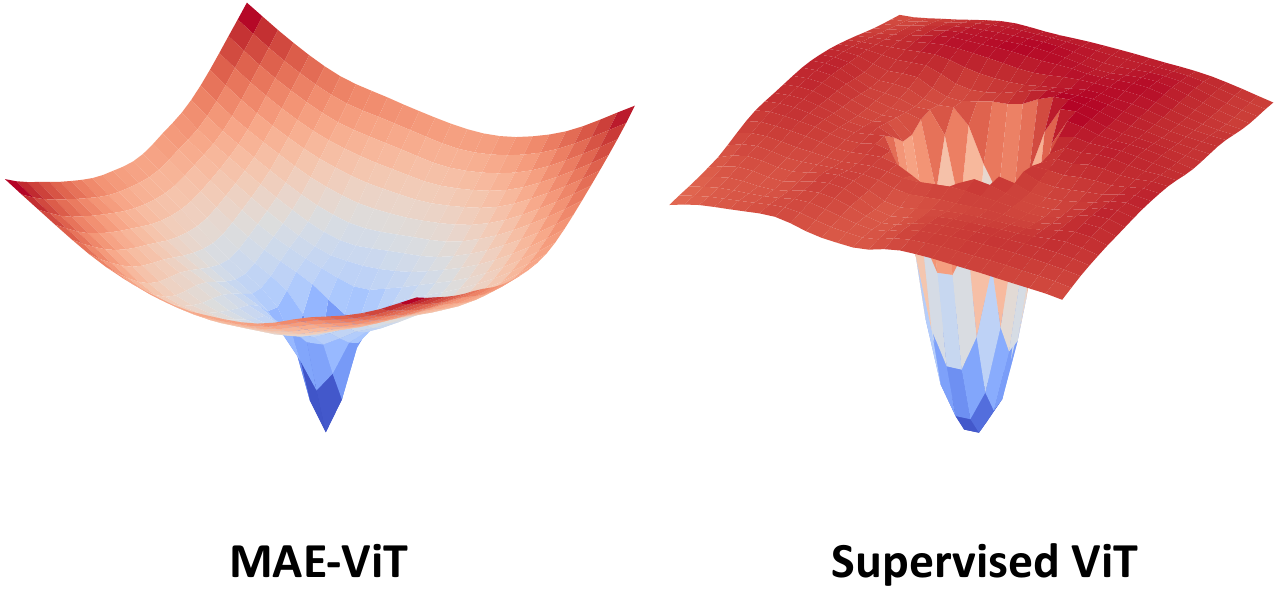}
\end{center}
\vspace{-0.1in}
\caption{\textbf{Comparison between self-supervised vision transformer~(ViT)~\cite{dosovitskiy2021vit} by MAE~\cite{he2022mae} and supervised ViT.}
\texttt{MAE-ViT} is drawn by the pre-training loss~(masked patch prediction) and \texttt{Supervised ViT} is obtained by supervised image classification from scratch. We use the ViT-Base model.
MAE-ViT converges at a \textit{smoother} and \textit{wider} convex region than the superivsed ViT.
}
\label{fig:mae_sup}
\vspace{-0.5cm}
\end{figure}

In this work, we analyze the reason for better generalization of the self-supervised ViT when trained by MAE and the effect of the gradient correction of RC-MAE from the perspective of optimization.
To do this, we visualize loss landscapes~\cite{li2018visualizing} of the self-supervised ViTs as shown in~\Cref{fig:mae_sup} (MAE) and compare them with the supervised ViT trained from scratch.
Specifically, we compute the pretraining losses~(\eg, masked patch reconstruction) in MAE~\cite{he2022mae} and RC-MAE~\cite{lee2023rcmae} and visualize the loss landscapes by using the filter normalization method~\cite{li2018visualizing}.

Previous works~\cite{keskar2017on,pmlr-v80-kleinberg18a,jastrzębski2018on,2018Smith,chaudhari2017entropysgd,chen2022when} relating optimization with the loss landscape have demonstrated that a \textit{flatter} region with smaller curvature correlates well with the generalization of neural networks.
\cite{chen2022when} investigated the loss landscape of a supervised ViT from scratch.
However, these works have analyzed convolutional neural networks or vision transformers in a fully supervised learning setting, \eg, the image-classification task with cross-entropy loss.
To the best of our knowledge, our work is the first study to visualize the loss geometry of \textit{self-supervised} vision transformers.

Through the lens of the loss landscape, we summarize our findings on the self-supervised ViT as follows:
\vspace{-0.1in}
\begin{itemize}
  \item Self-supervised ViTs have a \textit{flatter} and \textit{smoother} overall loss curvature and show better generalization properties compared to supervised ViT.
  \item The addition of an EMA-teacher~\cite{lee2023rcmae} allows MAE to widen the region of convexity, which leads to faster convergence.
\end{itemize}

In this work, we qualitatively analyze the optimized loss landscapes of the self-supervised ViT.
We leave more conclusive quantitative experiments for future works.
We hope that our work provokes further analysis into the self-supervised vision transformer from the perspective of optimization. 

\section{Preminimary: MAE and RC-MAE} \label{sec:pre}

\noindent
\textbf{The Masked Autoencoder}~(MAE)~\cite{he2022mae} performs self-supervised learning by randomly masking a large portion of input patches and then using a ViT-based encoder~$f$ and subsequently reconstructing the masked patches with decoder~$h$, given the encoded visible patches. 
MAE splits the input image $X\in\mathbb{R}^{C\times H \times W}$ into $N$ disjoint patches $\tilde{X}\in\mathbb{R}^{N \times(P^2\cdot C)}$ where $P^2$ represents the area of a patch. 
MAE then masks a random subset of $\mathbf{x}_i \in \tilde{X}, \; \forall \; i \in \mathcal{M}$, with $\mathcal{M}$ being the indices of the mask tokens. The visible patches, $\mathbf{x}_j \in \tilde{X}, \; \forall \; j \in \mathcal{V}$ (with $\mathcal{V}$ being the indices of the visible patches) are given to the encoder which produces an encoded vector $\boldsymbol{z}=f(\{\mathbf{x}_j: j \in V\})$. 
Subsequently, the decoder~$h$ performs a reconstruction of the set of masked patches $\{\mathbf{x}_i: i \in \mathcal{M}\}$ given the encoded visible patches, $\hat{Y} = h(\boldsymbol{z};\{\mathbf{x}_j: j \in M\})$, where $\hat{Y}\in\mathbb{R}^{N \times(P^2\cdot C)}$. 
The loss function $\mathcal{L}_r$ is then only computed on the masked patches with a mean squared error loss function.
\begin{equation}\label{eq:mae_loss}
    \loss{r} = \frac{1}{|\mathcal{M}|}\sum_{i \in \mathcal{M}} \big\Vert \tilde{X}_i - \hat{Y}_i \big\Vert^2_2.     
\end{equation}


\noindent
\textbf{The Reconstruction Consistent Masked Autoencoder}~(RC-MAE)~\cite{lee2023rcmae} provided analysis into a common technique utilized~\cite{grill2020byol,caron2021dino,zhou2021ibot,assran2022msn} throughout self-supervised learning, the EMA teacher~\cite{tarvainen2017mean}. An EMA teacher $T$ is composed of an exponential moving average of previous students $S$ with $\alpha \in [0, 1]$ by $T^{(t)} = \alpha T^{(t-1)} + (1 - \alpha)S^{(t)}$ which can be recursively expanded to $T^{(t)} = \sum_{i=0}^t \alpha^i (1 - \alpha)S^{t-i}$. 
In addition to the MAE's~\cite{he2022mae} reconstruction objective $\loss{r}$, the teacher provides a consistency target $\hat{Y}^\prime$ to the student network~(\eg, MAE). 
Thus, the student network is optimized by the objectives as follows: 
\begin{equation}\label{eq:rc-mae-s}
    \frac{1}{|\mathcal{M}|}\sum_{i \in \mathcal{M}}\color{black}{(\color{cobalt}{{
      \underbrace{\color{black}{\norm{\tilde{X}_i - \hat{Y}_i}^2}}_\text{reconstruction}}} 
      \color{black}{+} 
      \color{red}{\underbrace{\color{black}{\norm{\hat{Y}_i - \hat{Y}^\prime_i}^2}}_\text{consistency}}\color{black}{)}},
\end{equation}
It was revealed that in a simple linear model~\cite{lee2023rcmae}, the teacher acts in essence like a gradient memory which removes previous gradient directions conditionally if the current input $\mathbf{x}_i$ is similar to a previous input $\hat{\mathbf{x}}_j$ measured by the dot product. 
Likewise, when the current inputs are orthogonal to previous inputs, the dot product is $0$ and the teacher gives no corrective signal.
Thus, the EMA teacher may stabilize training by actively preventing overfitting when it is likely (\eg when there is low input diversity) and allowing the model to learn new knowledge when overfitting is less likely (\eg there is high input diversity).      

\vspace{+1.3cm}
\begin{wraptable}{r}{77mm}
\centering
\caption{\textbf{Downstream tasks results} using ViT-Base model. FN and LN respectively denote end-to-end finetuning and linear probing on ImageNet-1K. bbox/mask AP are results utilizing Mask R-CNN Benchmarking~\cite{li2021benchmarking}. These results demonstrate that MIM-based pre-training shows better \textbf{generalization} capability than fully supervised training from scratch.}
\begin{tabular}{lcccc}
\toprule
Method & FN            & LN            & bbox AP       & mask AP       \\ \midrule
Supervised              & 82.3         & -             & 47.9          & 49.3 \\
MAE~\cite{he2022mae}    & 83.4          & 67.3          & 50.3          & 44.9          \\
RC-MAE~\cite{lee2023rcmae} & \textbf{83.6} & \textbf{68.4} & \textbf{51.0} & \textbf{45.4} \\ \bottomrule
\end{tabular}
  \label{tab:result}%
\end{wraptable}

\section{Loss landscape}
To visualize loss landscapes of vision transformer, we follow the visualization strategy from~\cite{li2018visualizing} called filter-wise normalization.
Specifically, \cite{li2018visualizing} obtains two random Gaussian direction vectors $\delta$ and $\eta$ for each parameter $\theta$ to visualize the loss surface within the 2D projected space. 
Note that the direction vectors are normalized to have the same norm as the corresponding parameter $\theta$. 
And then, the loss surface is obtained by evaluating the loss on 2D points along the two directions as follows:
\begin{equation}
    f(\alpha, \beta)=L(\theta+\alpha\delta+\beta\eta),
\end{equation}
where $L$ is the loss function for a network parameterized by $\theta$. $\alpha$ and $\beta$ are varied scalar values from -1 to 1 corresponding to the $x$-axis and $y$-axis in the loss surface, respectively. please refer to~\Cref{sec:supple:Implementation Details} for more implementation details.



\subsection{Analysis}
\cite{li2018visualizing} demonstrates that the \textit{flatness} of minimizers correlates well with generalization due to this visualization method based on filter normalization. 
In addition, numerous studies~\cite{chaudhari2017entropysgd, keskar2017on,2018Smith,pmlr-v80-kleinberg18a,jastrzębski2018on,li2018visualizing,Zela2020Understanding,pmlr-v119-chen20f,chen2022when} have drawn a conclusion that {\textbf{neural networks tend to generalize better when they converge to a \textit{flat} region with small curvature and a wide region of convexity}.
Since we already confirmed that self-supervised learning methods, MAE~\cite{he2022mae} and RC-MAE~\cite{lee2023rcmae}, using vision transformers~\cite{dosovitskiy2021vit} generalize better than purely supervised learning in~\Cref{tab:result}, in this section, we analyze the reason for the better generalization capability of the self-supervised methods and investigate the effect of the EMA Teachers in RC-MAE from the \textit{optimization} perspective by observing the loss landscapes.

\begin{figure*}
\begin{center}
\includegraphics[width=1.\textwidth]{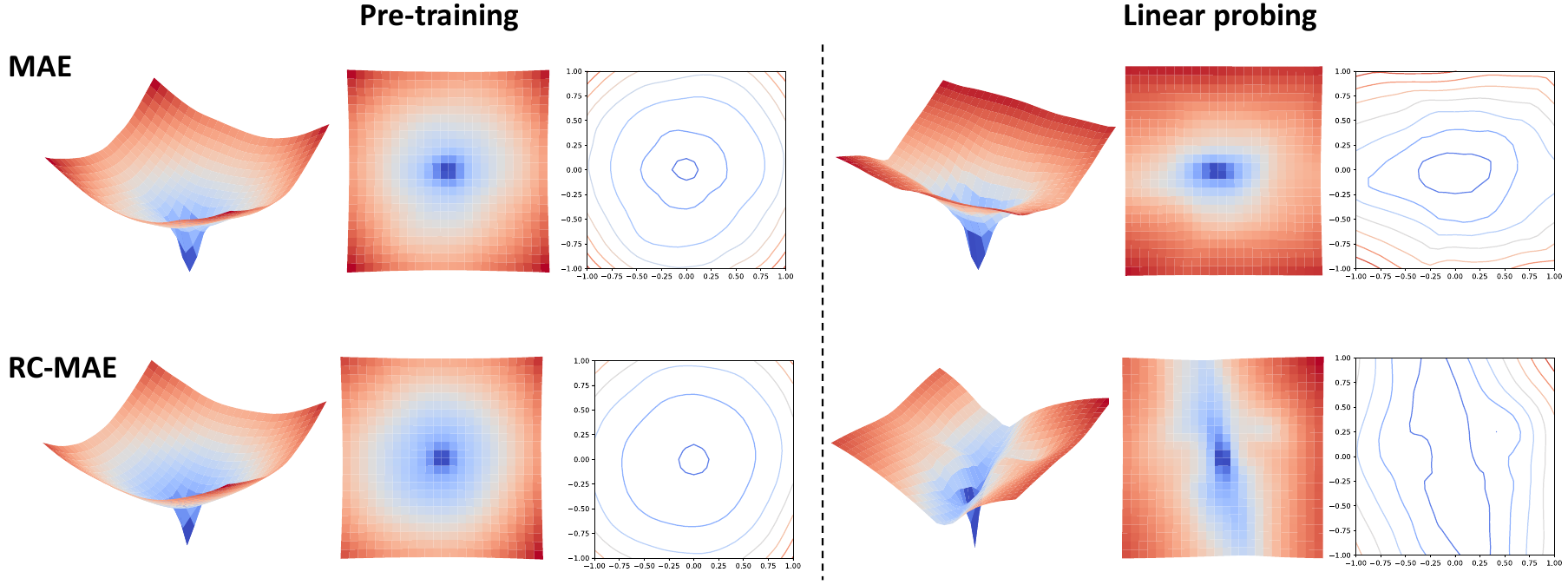}

\end{center}
\caption{
\textbf{Comparison between MAE~\cite{he2022mae} and RC-MAE~\cite{lee2023rcmae} using pre-training~(left) and linear probing~(right) weights.}
We use the ViT-B which is pre-trained for 1600 epochs with MAE or RC-MAE.
While The pre-training computes the masked reconstruction loss through mean-square error loss, the linear probing task computes image classification loss with cross-entropy loss. 
}
\vspace{-0.3cm}
\label{fig:pt_ln}
\end{figure*}

\noindent
\textbf{MAE-ViT vs. Supervised-ViT.}
As shown in~\Cref{fig:mae_sup}, the vision transformer~\cite{dosovitskiy2021vit} supervised from scratch
shows a narrower convergence region~(\ie, convex region) which is also
observed by~\cite{chen2022when}.
In contrast, we can observe that the self-supervised vision transformer by MAE~\cite{he2022mae} exhibits a much wider convex area of the loss landscape.
This demonstrates that the self-supervised methods likely converge under a broader set of initial conditions compared to fully supervised models.
We note that this \textit{smoother} and \textit{wider} loss landscape and better downstream task performance~(\ie, generalization result) are consistent with the fact that \textit{a wide convex region with a small curvature correlates well with the generalization of neural networks} in the optimization literature~\cite{keskar2017on,pmlr-v80-kleinberg18a,jastrzębski2018on,2018Smith,chaudhari2017entropysgd,chen2022when}.


\noindent
\textbf{MAE vs. RC-MAE.}
\Cref{fig:pt_ln} illustrates loss landscapes of the self-supervised vision transformers by MAE~\cite{he2022mae} and RC-MAE~\cite{lee2023rcmae} which result from pre-training and linear probing.
For the pre-training as shown in~\Cref{fig:pt_ln}~(left), the top-view and 2D loss contour of the landscapes show that RC-MAE converges from a \textit{wider} region of convexity than MAE.
Additionally, as shown in~\Cref{fig:pt_ln}~(right), linear probing results of both MAE and RC-MAE have more complex loss curvature than that of the pre-training loss.
We speculate that the linear probing task freezing the feature weights and only learning a linear layer is hard to optimize for classifying 1K categories.
Similar to the pre-training result, RC-MAE has a wider convex region than MAE.
As the only difference between MAE and RC-MAE is the addition of an EMA teacher, this suggests that this wider convex region could be attributed to the effect of the gradient correction by the EMA-teacher in RC-MAE.
Furthermore, Lee et al.~\cite{lee2023rcmae} have demonstrated that the convergence speed of RC-MAE is faster than MAE by comparing reconstruction loss graphs and finetuning accuracies in their paper. 
These loss landscape comparisons between MAE and RC-MAE together with the experiments in \cite{lee2023rcmae} support the better convergence properties of RC-MAE.



\section{Conclusion and Future works}
In this work, we have investigated the reason for the generalization capability of the self-supervised vision transformer and the gradient correction effect of RC-MAE by visualizing the loss landscapes of various self-supervised ViT's.
Through the lens of loss landscapes, we have observed interesting things: (1) Self-supervised  vision transformers have a smoother and wider overall loss curvature than fully supervised ViT's. 
(2) The self-distillation architecture~(\ie, EMA-teacher) allows MAE to widen the region of convexity, accelerating convergence speed.
However, there is still room for further exploring the effect of self-supervised learning.
\textbf{Batch size}:\cite{li2018visualizing} shows that batch size affects the sharpness of the minimizer.
Future works could explore the effect of the batch size on MIM pre-training.
\textbf{Optimizers}: \cite{chen2022when} utilize a sharpness-aware optimizer which could have some of the same properties as the EMA teacher in RC-MAE. A direct comparison and analysis could yield some interesting takeaways. 
\textbf{Comparison with other self-supervised methods}: We have analyzed only MIM-based self-supervised methods~\cite{he2022mae,lee2023rcmae}. However, before the emergence of MIM methods, instance discrimination tasks~(\eg, contrastive learning)-based methods~\cite{he2020moco,chen2020mocov2,chen2021mocov3,chen2020simclr,grill2020byol,chen2020simclr2,caron2021dino} had been dominant. Therefore, it would be interesting to investigate and compare the non-MIM methods from the standpoint of loss geometry.
\textbf{Quantitative analysis}: We have performed only qualitative analyses using loss landscapes. 
However, \cite{chen2022when} quantifying the \textit{average flatness} and the degree of loss \textit{curvature} by calculating the training error under Gaussian perturbations on the model parameter and the dominant Hessian eigenvalue, respectively.
Thus, we look forward to future works which may quantitatively compare those metrics of optimization dynamics.

\section{Acknowledgement}  

This work was partly supported by Institute of Information \& Communications Technology Planning \& Evaluation(IITP) grant funded by the Korea government(MSIT) (No. RS-2022-00187238, Development of Large Korean Language Model Technology for Efficient Pre-training~(80\%) and No. 2020-0-00004, Development of Previsional Intelligence based on Long-term Visual Memory Network~(20\%)).


\bibliographystyle{plain}
\bibliography{main}

\clearpage
\appendix

\paragraph{\LARGE{Appendix}}

\section{Implementation Details \label{sec:supple:Implementation Details}}

\noindent
\textbf{Setup.}
For reducing computation cost, we compute the loss landscape on the ImageNetV2~\cite{recht2019imagenetv2} validation set which has 10K images.
We use the pre-trained weight for the supervised ViT the original authors provided via the \texttt{timm}~\cite{rw2019timm} library.
As existing loss landscape works~\cite{li2018visualizing,chen2022when,shen2022sliced} visualize neural networks by using a supervised image classification task with cross-entropy loss, we also visualize the loss landscape of the supervised ViT by computing image classification loss.

\noindent
\textbf{Pre-training.}
Unlike the supervised ViT using image classification task, MAE~\cite{he2022mae} and RC-MAE~\cite{lee2023rcmae} perform a masked patch reconstruction task for pre-training.
Since we want to analyze the loss dynamics during the pre-training phase, we perform the filter normalization method~\cite{li2018visualizing} with the masked patch reconstruction loss, instead of a classification loss.
Thus, we visualize the loss landscapes of MAE and RC-MAE by using~\Cref{eq:mae_loss} and~\Cref{eq:rc-mae-s}, respectively.
To do this, firstly, we pre-train both MAE and RC-MAE with vision transformer~(ViT~\cite{dosovitskiy2021vit} base) for 1600 epochs in a 1-node server equipped with 8 GPUs, following the setup of RC-MAE.
The visualizations of loss landscapes are then created using the pre-trained weights.

\noindent
\textbf{Linear probing.} 
Linear probing or fine-tuning on downstream tasks have been used as standard protocols for self-supervised learning~\cite{he2020moco,chen2020simclr,grill2020byol,chen2020mocov2,caron2020swav,chen2021mocov3,caron2021dino,he2022mae,zhou2021ibot,assran2022msn}.
Linear probing freezes the pre-trained weights~(\ie, feature) and learns a linear classifier on top of the frozen features in the ImageNet classification task.
We assume that this linear probing could also prove to be an insightful task in analyzing the pre-training because the pre-trained features are frozen.
To do this, we perform the linear probing evaluation using the pre-trained models for 1600 epochs by MAE and RC-MAE and achieve the results as shown in~\Cref{tab:result}.
We obtain loss landscapes from the linear probing weights of both MAE and RC-MAE by using cross-entropy loss.

\noindent
\textbf{Downstream tasks.}
To validate the generalization capability of the vision transformer which is pre-trained by MAE and RC-MAE, we perform downstream tasks including end-to-end ImageNet fine-tuning, object detection, and instance segmentation using Mask R-CNN benchmarking~\cite{li2021benchmarking} on the COCO~\cite{lin2014coco} dataset.
As shown in~\Cref{tab:result}, MAE and RC-MAE consistently outperform the supervised model in all tasks, demonstrating that self-supervised methods show better generalization than supervised learning. 
Furthermore, RC-MAE also achieves higher performance than MAE in all downstream tasks.

\end{document}